\documentclass[conference,a4paper]{IEEEtran}
\IEEEoverridecommandlockouts
\usepackage{cite}
\usepackage{amsmath,amssymb,amsfonts}
\usepackage{algorithmic}
\usepackage{graphicx}
\usepackage{textcomp}
\usepackage{xcolor}
\usepackage{pdfpages}
\def\BibTeX{{\rm B\kern-.05em{\sc i\kern-.025em b}\kern-.08em
    T\kern-.1667em\lower.7ex\hbox{E}\kern-.125emX}}
\begin{document}

\makeatletter
\newcommand{\newlineauthors}{%
  \end{@IEEEauthorhalign}\hfill\mbox{}\par
  \mbox{}\hfill\begin{@IEEEauthorhalign}
}
\makeatother

\title{OmniVoxel: A Fast and Precise Reconstruction Method of Omnidirectional Neural Radiance Field
}

\author{\IEEEauthorblockN{ Qiaoge Li}
\IEEEauthorblockA{\textit{University of Tsukuba} \\
Tsukuba, Japan \\
li.qiaoge@image.iit.tsukuba.ac.jp}
\and
\IEEEauthorblockN{ Itsuki Ueda}
\IEEEauthorblockA{\textit{University of Tsukuba} \\
Tsukuba, Japan \\
ueda.itsuki@image.iit.tsukuba.ac.jp}
\and
\IEEEauthorblockN{Chun Xie}
\IEEEauthorblockA{\textit{University of Tsukuba} \\
Tsukuba, Japan \\
xiechun@ccs.tsukuba.ac.jp}
\newlineauthors
\IEEEauthorblockN{ Hidehiko Shishido}
\IEEEauthorblockA{\textit{University of Tsukuba} \\
Tsukuba, Japan \\
shishido@ccs.tsukuba.ac.jp}
\and
\IEEEauthorblockN{Itaru Kitahara}
\IEEEauthorblockA{\textit{University of Tsukuba} \\
Tsukuba, Japan \\
kitahara@iit.tsukuba.ac.jp}
}

\maketitle

\begin{abstract}
This paper proposes a method to reconstruct the neural radiance field with equirectangular omnidirectional images. Implicit neural scene representation with a radiance field can reconstruct the 3D shape of a scene continuously within a limited spatial area. However, training a fully implicit representation on commercial PC hardware requires a lot of time and computing resources (15 $\sim$ 20 hours per scene). Therefore, we propose a method to accelerate this process significantly (20 $\sim$ 40 minutes per scene). Instead of using a fully implicit representation of rays for radiance field reconstruction, we adopt feature voxels that contain density and color features in tensors. Considering omnidirectional equirectangular input and the camera layout, we use spherical voxelization for representation instead of cubic representation. Our voxelization method could balance the reconstruction quality of the inner scene and outer scene. In addition, we adopt the axis-aligned positional encoding method on the color features to increase the total image quality. Our method achieves satisfying empirical performance on synthetic datasets with random camera poses. Moreover, we test our method with real scenes which contain complex geometries and also achieve state-of-the-art performance. Our code and complete dataset will be released at the same time as the paper publication.
\end{abstract}

\begin{IEEEkeywords}
Human-centered computing, Virtual Reality, Immersive Experience, Free-viewpoint Videos, Image-based Rendering
\end{IEEEkeywords}

\section{Introduction}
With the increasingly accurate reproduction of natural scenes by neural rendering technology, lifelike reconstruction of real scenes in Virtual Reality (VR) is gradually becoming possible. The most representative techniques to promote this research are Neural Radiance Field (NeRF) and Multi-Plane Image (MPI). In the foreseeable future, real scene reconstruction will increasingly be applied to the field of VR as well as many other multimedia fields like tele-education and virtual tourism. However, most cameras can only provide perspective views as input, and the number of images required to build a complete scene can be huge. Using omnidirectional shots to reconstruct the entire scene reduces the need for the number of pictures and the extra attention to the coverage area that needs to be considered for the shot. 

Omnidirectional reproduction of real scenes has been put into use for several years. Years before, this procedure required the use of multiple perspective views for stitching. However, for hand-held photography, it is hard to strictly make the positions of different cameras identical when capturing multiple perspective images, which leads to the result that the stitching will produce displacements and distortions at the seams of adjacent images. Today, omnidirectional scenes are shot with ultra-wide-angle fish-eye lenses and built-in algorithms for real-time stitching. The quality of the omnidirectional images generated in this way is stable. 

Our goal is to reconstruct the 3D space of the captured scene holistically with equirectangular omnidirectioal inputs in a relatively short time. We focus on the fact that the omnidirectional image contains the ray information from all directions space to the camera position. Therefore, from the original hypothesis of NeRF, we assume that learning ray information from multiple spherical panoramas can generate a continuous omnidirectional radiance field that is fully capable of representing visual information within a whole space. However, training on new scenes based on this assumption takes a vast amount of time, also the limitations of NeRF in representing rays lead to a low quality reconstruction. Our proposed method modeled the 3D scene into latent voxels to accelerate the reconstruction speed for the radiance field and to increase the speed. When modeling rays from multi-view panoramas with original assumptions of NeRF, we find that the intersection among rays is not evenly distributed from the center to the edge of the scene. This property results in an uneven quality of the reconstruction of the scene from inside to outside for unbounded scenes. Therefore,we propose to use the spherical coordinate voxelization method instead of the traditional cubic voxel representation using Cartesian coordinates. This paper discusses the reconstruction quality as well as the processing speed of previous Omnidirectional NeRF and the two voxelization with the proposed method for different scenarios. 

Our voxelization method adopt a tensor decomposition approach to reduce spatial complexity, enabling us to train models with higher resolution. Due to the much higher frequency of omnidirectional images than perspective ones, the traditional positional encoding method is not enough to get satisfying results. We applied the axis-aligned positional encoding method to the color features to increase the detail quality for the final results on captured images. In addition, we produced a complete dataset that can be used for indoor reconstruction tasks, including 315 equirectangular photographs captured by a high-resolution omnidirectional camera under fixed lighting conditions and their camera parameters. We also provide ground truth depth information scanned by the LiDAR scanner. In summary, our main contributions are listed as follows:

\begin{itemize}
\item We propose a method that uses only RGB information from multiple captured panoramas to reconstruct the radiance field holistically within a short time.

\item We elaborate and experimentally demonstrate that the reconstruction quality of the voxel-based partial explicit representation is better than the ray-based implicit representation when using panoramas for free viewpoint image generation.



\item We provide a comprehensive dataset for omnidirectional novel view synthesis task including over 1500 equirectangular omnidirectional photographs with their camera parameters. This dataset contains four different scenes including indoor, outdoor and synthesized scenes. We also provide the ground truth depth information for indoor and synthesized scenes.
\end{itemize}

\section{Related Work}
Novel view synthesis aims to solve this problem by synthesizing new views using a limited number of RGB images.  In the past three years, various implicit neural scene representation methods using deep learning have achieved compelling results for the novel view synthesizing task. \cite{NIR3_llff,NIR6_imap,NIR7_nerf,NIR8_ibrnet}. 
Among them, Neural Radiance Field (NeRF) \cite{NIR7_nerf} and its derivative methods \cite{nerf5_mip,mipnerf360} receive wide range of attention. Unlike traditional scene reconstruction
\cite{IBR1,MBR1_kanade,MPI1_stereo}, which requires an explicit representation of the scene geometry as a first step, NeRF implicitly represents the scene as rays observed from the viewpoint using neural networks with the structure of multilayer perceptron (MLP).
This representation allows the reconstruction of the scene to be continuous within space. Another essential feature of NeRF (i.e. ``positional encoding") enables low-dimensional inputs to be projected into higher dimensions\cite{NIR7_nerf}, which not only allows the gradient of the interpolation function describing the volume density of the space to be calculated but also significantly improves the accuracy of the final function obtained by deep learning.

For boundless scenes, however, the performance of NeRF is relatively low. The intersection of light sampled from different cameras becomes increasingly sparse as the actual spatial distance gets farther away, which reduces the quality of NeRF when learning the representation of a distant scene or the objects in the background. NeRF++\cite{nerf++} proposes to solve this problem by sampling based on disparity outside a specific range instead of sampling based on distance. 

Another common problem that NeRF ignored is that the pixel on each image cannot actually be represented by the rays it modeled. This produces ambiguous features sampled on adjacent space points, thereby limiting NeRF’s performance significantly, especially for unbounded scenes. Mip-NeRF-360\cite{mipnerf360} adopted cone sampling and integrated positional encoding to substantially increase the scene representation quality for unbounded scenes.

With the gradual unification of the structure of omnidirectional cameras, i.e., consisting of multiple wide-angle lenses, and the gradual decrease of their cost, the acquisition of spherical panoramic photos is no longer complicated. Therefore, research on synthesizing novel views directly on omnidirectional images comes out these years \cite{OFV_journal,OmniPhotos}. Recently, some contemporary works have adopted equirectangular omnidirectional images as input for the reconstruction of the radiance field. Several recent works \cite{omni-nerf} show their attempt at omnidirectional radiance field representation. In their methods, the depth information of the scene is known, which allows them to obtain accurate results with almost no prediction of the transparency of the scene.  Another work \cite{omni_nerf_old}used only RGB images and the camera parameters as input for reconstruction, but their results are very bad compared with SOTA methods. One common problem is that their methods take a lot of time ($15\sim 20$ hours per scene) for reconstruction, while our method only requires a much shorter time ($20\sim 40$ minutes per scene).  

\section{Method}

\subsection{Flexible Spherical Voxelization}

Since it is impossible to choose the shooting direction with an omnidirectional camera, a spherical panorama is thus more likely to contain a boundless scene than a perspective view. In addition, commercial omnidirectional cameras require handheld or tripod shots, which will result in a significant portion of the panorama being occupied by near objects. These two points lead us to consider the reconstruction quality of both near and far scenes. If we use voxels uniformly distributed along the Cartesian coordinate system, then as the distance to the camera increases, less and less light will pass through the same voxel. As a result, the quality of the reconstruction will vary with the distance to the camera. Therefore, we devise a spherical voxelization approach to balance the reconstruction quality of distant and close views for unbounded scenes.

In the proposed method, the spherical voxel representation explicitly models the color and density features in spherical grid cells. We store these modalities separately in 3 tensors along $r$, $\theta$, and $\phi$ coordinates. We optimize the partitioning on the $r$-axis in the voxelization process. The interval of voxelization along the $r$-axis decreases as the distance from the center increases as shown in the following equation:
\begin{equation}
    T_i = \ln{(t_i+1)} ,  i \in [0,N],
    \label{eqn:3}
\end{equation}
where $T_i$ is the real sampling distance on the radius for $i$th voxel, and $t_i$ represents the distance between the $i$th evenly distributed voxel and the ray origin. $N$ is the total number of voxels sampled on the radius. This distribution effectively reduces the difference in the number of voxels per ray passed by cameras farther away from the scene center during sampling relative to cameras closer to the scene center. Then, the scene would be easily represented by interpolation. The trilinear interpolation method has been adopted to interpolate the queried 3D positions.

\begin{figure*}[t]
  \centering
   \includegraphics[width=\linewidth]{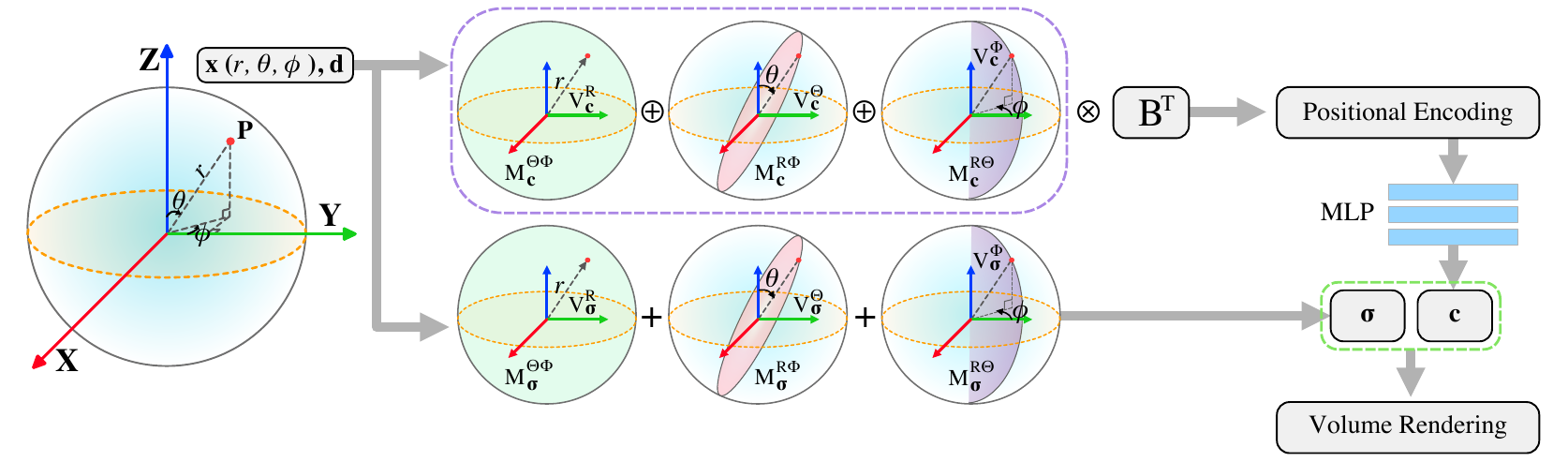}
   \caption{Tensor decomposition and rendering: We decompose the tensors representing spherical voxelized feature grids into a set of vectors($V$) and matrices ($M$). Same as equation(5) the appearance features are multiplied by an appearance matrix B. The volume density values are directly summed by the density components, while the appearance features are proceeded by axis aligned positional encoding, and finally put into MLP for RGB regression. Then RGB values of image can be rendered by volume density $\sigma$ and appearance feature $c$.  }
   \label{fig:7}
\end{figure*}

\subsection{Tensor Decomposition.} 
Inspired by TensoRF~\cite{TensoRF}, we applied Vector-Matrix (VM) decomposition on the tensors to decrease memory consumption. In our case, scene is represented by 3D tensor modes corresponding with $r$,$\theta$, and $\phi$ axis. Given a 3D tensor $T \in \mathbb{R}^{R\times \Theta\times \Phi}$, VM decomposition factorizes a tensor into multiple vectors and matrices; the following equation shows this process:

\begin{equation}
    T = \sum_{n=1}^{N_1} V_n^{R} \otimes M_n^{\Theta,\Phi} + \sum_{n=1}^{N_2} V_n^{\Theta} \otimes M_n^{R,\Phi} + \sum_{n=1}^{N_3} V_n^{\Phi} \otimes M_n^{R,\Theta},
    \label{eqn:vm}
\end{equation}

where $V_n^{R}$,  $V_n^{\Theta}$, $V_n^{\Phi}$ corresponds to a rank-one tensor component, $M_n^{\Theta,\Phi}$, $M_n^{R,\Phi}$, $M_n^{R,\Theta}$, are matrix factors for two (represented by superscripts) of the three modes that different from the tensor components denoted in the corresponding vector. In 3D representation with the Cartesian coordinate, a scene can distribute and appear equally complex along its three axes, and in that case, N1 = N2= N3. In our case, we also set N1 = N2 = N3, the balance between $R$ (N1) and $\Theta,\Phi$ (N2, N3) is adjusted by the scale of $R$. 3D tensor would be enough for representing the volume density, while color requires one more dimension for the representation of channels. This is represented by a vector $b$ multiplied by each color tensor. In addition, we use three component tensors to simplify notation and the following discussion in later sections: $A^R_n=V_n^R \otimes M_n^{\Theta\Phi}$, $A^\Theta_n=V_n^\Theta \otimes M_n^{R\Phi}$, and $A^\Phi_n=V_n^\Phi \otimes M_n^{R\Theta}$. Then the volume density and color of the 3D voxels can be expressed as
\begin{equation}
\label{eq:4}
\left\{
\begin{aligned}
    & G_{\sigma} = \sum_{n=1}^{N_\sigma} \sum_{m\in R\Theta\Phi} A^m_{\sigma, n}, \\
    & G_{c} =\sum_{n=1}^{N_c} A_{c,n}^R \otimes b_{3n-2} +A_{c,n}^\Theta \otimes b_{3n-1} +A_{c,n}^\Phi \otimes b_{3n},
\end{aligned}
\right.
\end{equation}

where $G_\sigma$ and $G_c$ represents the 3D geometric tensor for density and color. In total, we parameterize the voxels into  $3N_\sigma+3N_c$ matrices and $3N_\sigma+6N_c$ vectors. 
Fig.~\ref{fig:7} gives a more intuitive explanation of the whole procedure. An important benefit of representing voxels using tensor decomposition is that the computational effort of trilinear interpolation necessary to reconstruct neural radiation sites is greatly reduced. Interpolate the component tensor trilinearly is equivalent to interpolate the corresponding modes of its vector/matrix factors linearly/bilinearly. Therefore, it save a lot of time and computing costs which enables us to train higher resolution of images and voxels than other volelization approaches.

\subsection{Positional Encoding}
In original NeRF, positional encoding is essential for performance improvement.
It is attributed to the difficulty of MLPs to learn high-frequency functions due to spectral biases, which can make the network learned by MLPs with only 5D inputs unable to restore scene details adequately. The NeRF experiments obtained good results with a heuristic sinusoidal mapping of the input coordinates (i.e., ``position encoding") to allow MLPs to represent higher frequency content. Since the average amount of information per pixel contained in an equirectangular panorama is higher than that of a perspective view with the same number of pixels in most cases, the frequency of information required to be restored for our task is much higher than that of the original NeRF. We applied positional encoding along the direction of aligned axes to improve the reconstruction quality of high-frequency information. Our generic positional encoding mapping $\gamma$ inputs points $\bf{v} \in [0,1)^d$ to the surface of a hypersphere that has much higher dimensions with a set of sinusoids:

\begin{equation}
\label{eq:5}
\begin{aligned}
\gamma(\bf{v})=[a_1\cos(2\pi \bf{b}_1^T \bf{v}),a_1\sin(2\pi \bf{b}_1^T \bf{v}),\\ \cdots, \bf{a_m}\cos(2\pi \bf{b}_m^T \bf{v}),a_m\sin(2\pi \bf{b}_m^T \bf{v})]^T.
\end{aligned}
\end{equation}

For our axis aligned positional encoding:

\begin{equation}
\label{eq:10}
\bf{a_i} = J^d , \bf{b_i} = \sigma^{j/m},
\end{equation}

where  $j$ = $0,\cdots,m-1$. $\sigma$ is a hyperparameter that is different for various tasks. In our case, $\sigma = 2$. In both positional encoding methods, $a_i$ is a vector that only contains 1s with the same element number of the input dimension $d$. The embedding size for the positional encoding method is $m$. In our method, we applied the axis aligned positional encoding methods on the color features before input them to the MLP decoder. 
\subsection{Rendering and Learning} 
We render the image with volumetric differentiable renderer same as NeRF, for each pixel, the color result is integrated numerically by weighting the sum of the RGB values and the volume density at a set of discrete sampling points on the ray as the following equation:

\begin{equation}
\label{eq:6}
\hat{C}(\bf{r}) =\sum_{i=1}^{N}T_i(1-\exp(-\sigma_i\delta_i))\bf{c}_i, 
\end{equation}

\begin{equation}
\label{eq:7}
T_i = \exp(-\sum_{j=1}^{i-1}\sigma_i\delta_i),
\end{equation}
where $\bf{N}$ is the number of the sampling points,  $\bf\delta_i$ is the distance between adjacent samples. The function adopts traditional alpha blending method with alpha values $\alpha_i=1-\exp(-\sigma_i\delta_i)$ . $\delta_i$ represents the volume density at the sampled point. The $T_i$ function calculates the accumulated transmittance between the two samples along the ray.


With the above rendering method, we can render the omnidirectional image spherically, then we compare the image sphere and the ground truth and calculate the photometric loss and L1 sparsity loss. Our total loss function is like the following:

\begin{equation}
\begin{aligned}
    L = \|C - \tilde{C} \|^2_2 + \omega  \cdot L_1,
\label{eq:loss}
\end{aligned}
\end{equation}
where $\omega$ represents the weight for $L_1$ loss, and $\tilde{C}$ is the ground truth color.

\section{Experiments}

\begin{table}[]
\caption{Quantitative Results}
\label{tab:1}
\resizebox{0.47\textwidth}{!}{
\begin{tabular}{l|llll}
\hline
Metric\textbackslash Scene               & Indoor         & Outdoor1       & Outdoor2       & Blender        \\ \hline
PSNR$\uparrow$      &                &                &                &                \\ \hline
Omninerf            & 18.77          & 21.80          & 23.39          & 26.50          \\
Omninerf APE        & 19.73          & 22.32          & 23.79          & 27.01          \\
Omnivoxel Cubic     & 22.33          & 27.24          & 27.91          & 33.14          \\
Omnivoxel Cubic APE & 26.70          & \textbf{27.51} & \textbf{28.00} & 32.96          \\
Omnivoxel Sphere    & 23.38          & 27.10          & 27.95          & \textbf{33.23} \\
Omnivoxel Sphere APE& \textbf{26.87} & 27.38          & 27.94          & 33.19          \\ \hline
SSIM$\uparrow$      &                &                &                &                \\ \hline
Omninerf            & 0.752          & 0.764          & 0.815          & 0.902          \\
Omninerf APE        & 0.729          & 0.780          & 0.826          & 0.916          \\
Omnivoxel Cubic     & 0.805          & \textbf{0.824} & \textbf{0.893} & 0.936          \\
Omnivoxel Cubic APE & \textbf{0.815} & 0.808          & 0.892          & 0.932          \\
Omnivoxel Sphere    & 0.796          & 0.798          & 0.892          & \textbf{0.937} \\
Omnivoxel Sphere APE& 0.805          & 0.802          & 0.891          & 0.936          \\ \hline
LPIPS (Alex)$\downarrow$&            &                &                &                \\ \hline
Omninerf            & 0.583          & 0.376          & 0.346          & 0.183          \\
Omninerf APE        & 0.448          & 0.364          & 0.328          & 0.131          \\
Omnivoxel Cubic     & 0.301          & 0.355          & 0.234          & 0.125          \\
Omnivoxel Cubic APE & \textbf{0.266} & 0.344          & \textbf{0.233} & 0.116          \\
Omnivoxel Sphere    & 0.303          & 0.350          & 0.237          & \textbf{0.110} \\
Omnivoxel Sphere APE& 0.279          & \textbf{0.328} & 0.234          & 0.112          \\ \hline
LPIPS(VGG)$\downarrow$&              &                &                &                \\ \hline
Omninerf            & 0.538          & 0.418          & 0.407          & 0.373         \\
Omninerf APE        & 0.420          & 0.396          & 0.375          & 0.306         \\
Omnivoxel Cubic     & 0.339          & 0.379          & \textbf{0.319} & 0.294          \\
Omnivoxel Cubic APE & \textbf{0.329} & 0.369          & 0.327          & 0.303          \\
Omnivoxel Sphere    & 0.349          & 0.376          & 0.326          & \textbf{0.289} \\
Omnivoxel Sphere APE& 0.341          & \textbf{0.361} & 0.323          & 0.291          \\ \hline
\end{tabular}}
\end{table}

\subsection{Datasets}
We tested our method on the dataset made by our own. For synthesized data, we acquire camera parameters from Blender during rendering. For actual data, we use Pix4D mapper software to get the extrinsic parameters of the cameras. We use about 1/3 of them for training, 1/3 for validation, and 1/3 for testing. 

\begin{figure*}[t]
  \centering
   \includegraphics[width=\linewidth]{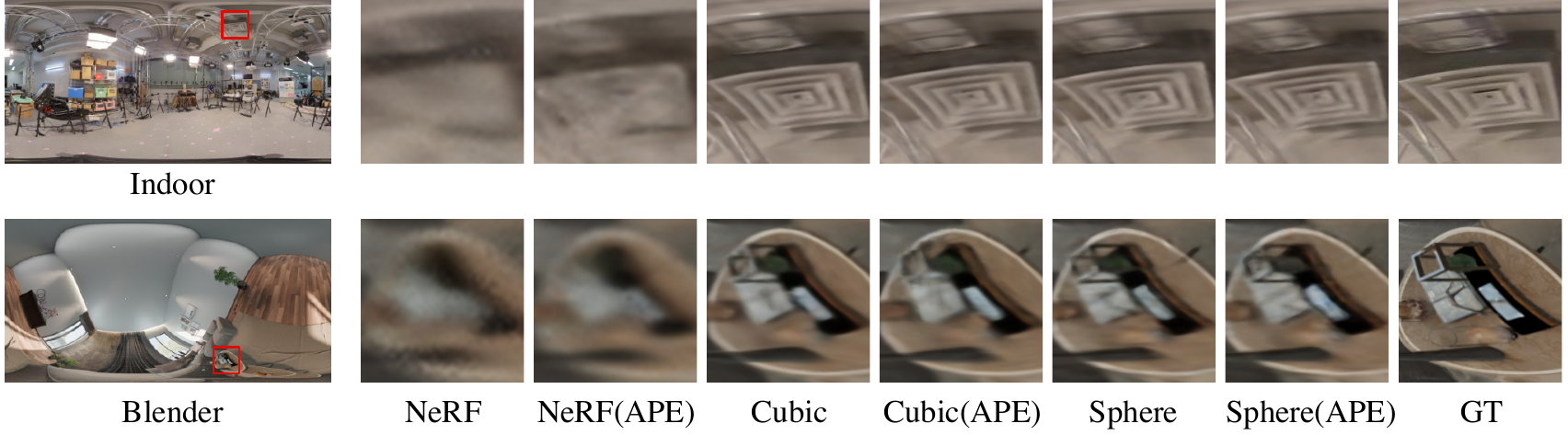}
   \caption{Close-up view of Results with different methods in different scenes, in which APE represents axis-aligned positional encoding, Cubic and Sphere means cubic and spherical voxelization methods, respectively.}
   \label{fig:8}
\end{figure*}

\subsection{Implementation Details}
During the experiment, the indoor and Blender images were resized to the resolution of $1024 \times 2048$ while the outdoor images kept their primitive resolution at $960 \times 1920$. We set the batch size 4096, which is the number of rays we sampled for each unit sphere. AdamW optimizer is used during the training process. We initialize the learning rate at $5\times10^{-4}$ and exponentially reduce it to $5\times10^{-5}$ during all training steps. We set the weight of $L_1$ loss as $8\times10^{-5}$ We train our voxelization models for $30$ thousand epochs in all experiments on a single RTX 3090 GPU. As a comparison, we trained the omnidirectional NeRF model for $250$ thousand epochs for their best performance.

\subsection{Results}

In Tab.~\ref{tab:1}, we compare different encoding methods within our method quantitatively in PSNR, SSIM, and LPIPS. Our experimental results in Fig.~\ref{fig:8} show that voxel representation achieved much better result than NeRF-based approach. The performance of spherical voxelization and cubic voxelization method are similar many different evaluation metrics. Axis-aligned positional encoding method has advantages in scenes with complex colors. Axis-aligned positional encoding performs well in real scenes because our reconstructed data are real-world 3D voxels, so the frequency domain of the information is oriented along the spatial axes (for both Cartesian and Spherical coordinate systems), rather than isotropic within dimensions. Even though we cannot compare directly with other methods on perspective dataset, it was evident that our result shows that the proposed method reached state-of-the-art performance for unbounded scene reconstruction and has similar performance to what mip-NeRF-360\cite{mipnerf360} and NeRF++ \cite{nerf++} have for their dataset. 

Due to space limitations, we cannot show the full-size equirectangular image results in the main text. They are included in the supplementary materials. As a result, spherical voxelization could balance the quality of the close and distant views of the center of the space. Also, the axis-aligned positional encoding method can reconstruct details of the tripod object while the original positional encoding can't. We also developed some satisfying flying-through videos with the trained models attached in the supplementary materials.

Another notable point is that our approach is far faster than directly applying the NeRF model to train on the spherical ray representation. Our method takes only 40 minutes to train a full representation on the Indoor dataset while NeRF takes more than 15 hours. Our proposed method makes it possible to reconstruct the entire scene using omnidirectional photos quickly and in high quality. 


\section{Conclusion}

We present a method for fast holistic reconstruction of the neural radiance field with multiple omnidirectional images. Our key idea is to use voxel grid representation and tensor decomposition to replace the fully implicit representation. We use the Unit Sphere model to sample the rays in different directions and adopt a spherical voxelization method to balance the quality of closer and distant views from the center of the scene. By modifying the positional encoding approaches, we quantitatively increase the quality of our result. Our method achieves satisfying empirical performance on synthetic datasets with random camera poses. Moreover,  our experiments on real datasets show that we can continuously reconstruct the unbounded omnidirectional scene at state-of-art-performance.
\bibliographystyle{IEEEtran}
\bibliography{IEEEabrv,IEEEexample}

\pagebreak



\section*{Supplementary material}

# OmniVoxel: A Fast and Precise Reconstruction Method of Omnidirectional Neural Radiance Field: Supplemental Materials

## I. Material Contents

In this supplemental material, we add some more related works, more figure for the procedure of the whole method and explanations of difference of the ray sampling error between the perspective camera and the omnidirectional camera. Also, we provide all the complete images used in the main text to compare the experimental results. We have attached some flying-through videos for reference, we recommend using the basic application of RICHO THETA (https://support.theta360.com/en/download/) for viewing them.

## II. Additional Related Works

### A. Voxel Based Radiance Field Representations

Fully implicit scene representations have the advantage of a high upper limit of reconstruction quality, but the time required for a scene to be trained from scratch is very long. In contrast, partially explicit reconstruction of neural radiation fields based on voxel rendering can substantially increase the training speed with similar quality of results. This approach has been a very active research direction in the last year [1]–[5]. However, these methods all use cubic voxels to represent the scene, which works well enough for datasets consisting of synthetic images or forward-facing photographs. However, for unbounded scenes, the use of uniformly distributed voxels may result in voxels far from the center of the captured scene not effectively representing the pixels representing the location in the image. This problem is similar to the one mentioned in the previous paragraph.

## III. Method Preliminaries

Our method aims to reconstruct the whole scene with only RGB information from multiple omnidirectional images captured in the space. We consider $\mathbf{o} \in \mathbb{R}^3$ as the position of a viewpoint, $\mathbf{d} \in \mathbb{R}^2$ as a vector representing a ray's direction, and $t \in \mathbb{R}_+$ as the distance of a point on the ray to the viewpoint. Therefore a discrete sample on the ray could be written as $\mathbf{r}(t) = \mathbf{o} + t\mathbf{d}$. Simultaneously, the whole scene is divided into $\mathbf{N}$ different voxels, which are arranged along with the spherical coordinates with the origin at the center of the scene inferred from the camera positions. We store the density $\sigma$ and color feature $\hat{C}$ information inside those grids as feature tensors and apply the tensor decomposition method to them. With the input of viewpoint position within the capturing scene $(x, y, z)$ and the viewing direction $(\theta, \phi)$, we integrate through these voxels along the ray directions using the neural volume renderer to finally obtain the RGB values. Then we can learn the view-dependent color features $\hat{C}(\mathbf{V})$ and the volume density $\sigma(\mathbf{V})$ of the volume in the whole space with very shallow MLPs. After performing an equirectangular projection on the inference results, we can obtain an omnidirectional image at the viewpoint.

## IV. Omnidirectional Ray Sampling

Existing radiance field reconstruction methods only sample on perspective images. Therefore, the sampling model is based on the pinhole camera model. The ray directions are determined by the intrinsic and extrinsic parameters of the camera. In the spatial coordinates of the actual scene, rays are sampled between the image plane and a parallel plane, which defines the farthest sampling distance as shown in Fig. 6(a). The commonly used exampled datasets are mainly rendered by the Blender software, and the backgrounds are set to transparent. This setup allows NeRFs to easily learn the transparency and the color features associated with the viewing directions of objects. However, in the case of forward-facing images, rays are usually converted to Normalized Device Coordinates (NDC) for sampling between the image plane and infinite far distance. NDC space can preserve parallel lines when converting the $z$ axis (camera axis) to be linear in disparity. However, 3D geometry in the NDC space is different from that in the real space. Thus, NeRF with NDC space has poor performance with the 360 datasets.

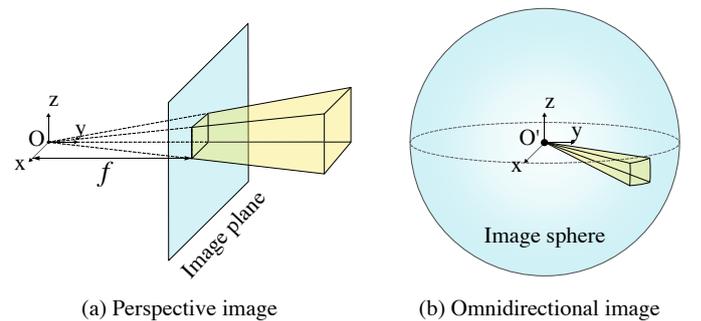

(a) Perspective image      (b) Omnidirectional image

Fig. 1. Comparison of ray sampling:(a) and (b) are ray sampling models for perspective and omnidirectional images. $O$ and $O'$ represents the camera positions. The yellow volumes on (a) and (b) are the effective sampling range in the real environment. $f$ represents the focal length for the pinhole camera model for perspective image, while in our model, the focal length of omnidirectional camera is zero.

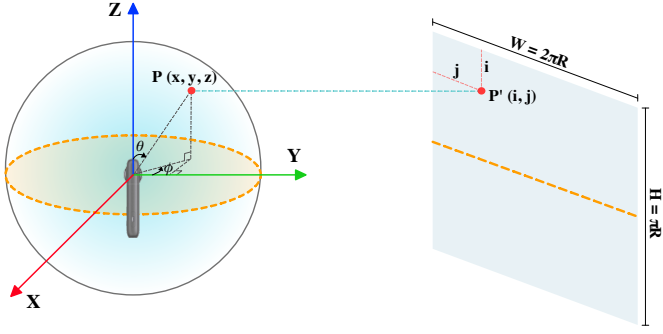

Fig. 2. Equirectangular projection: $P$ represents a point on the surface of the sphere, which is corresponding to $P'$, a point on the equirectangular image. The coordinates of $P'$ on the image $(i, j)$ are derived from the coordinates of $P(x, y, z)$ by Eq. 1, In the figure, $R$ represents the radius in the sphere and a scalar in the equirectangular image, which is related to the overall resolution. $\theta$ and $\phi$ represent the polar and azimuth angle of $P$ in spherical coordinate ,they are intermediate parameters for calculation.

On the other hand, omnidirectional camera doesn't have intrinsic parameters, therefore, rays can't be sampled with previously mentioned approaches. Our method samples the rays evenly in all directions with a unit sphere model for omnidirectional cameras. The focal length of an omnidirectional camera is zero, which means the representation of rays includes all the range from the camera position to the outer bound of sampling as shown in Fig. 6(b).

## V. Unit Sphere Camera Model

We project the equirectangular image to the two-step unit sphere model for sampling. First, we convert the 2D coordinate of the pixel on the equirectangular image to the normalized Cartesian coordinate of points on the unit sphere as the following equations:

$$\begin{cases} \theta = \dfrac{P'_j * \pi}{H}, \quad \phi = -\dfrac{P'_i * \pi}{H}, \\ P_x = \sin\phi * sin\theta, \quad P_y = \cos\phi, \quad P_z = -\sin\phi * \cos\theta, \end{cases} \quad (1)$$

where $P'_i$, $P'_j$ represent the row and column index of the equirectangular images, $H$ and $W$ represent its height and width.$\theta$ and $\phi$ represent the azimuth and polar angle for the point in spherical coordinate. $P_x$, $P_y$, $P_z$ are the position of the point in the right-hand Cartesian coordinate system. Fig. 2 shows the equirectangular projection from the sphere's surface to the image plane.

Our second step is to move and rotate the sphere with the camera's extrinsic parameters. In our proposed method, the camera orientation is derived from the Structure from Motion(SfM), as shown in Fig. 3(a),(b). We assume that when the camera rotation angle of all three axes is zero, the camera's orientation is nadir (looking down perpendicular to the ground)—the top of the equirectangular image points in the positive direction parallel to $Y$ axis.

$$\mathbf{X}' = \mathbf{R} \cdot (\mathbf{X} - \mathbf{T}) \quad (2)$$

Note that $\mathbf{X}$ and $\mathbf{X}'$ represent the coordinate of the points on the sphere before and after the conversion, respectively. $\mathbf{T}$ stands for the spatial coordinates of the camera position. $\mathbf{R}$ is the rotation matrix of the camera in the right-hand Cartesian coordinate system. After conversion, the image spheres will be aligned to same directions as Fig. 3(c).

In particular, since NDC space is unavailable in our method, we decide not to perform camera rays queries for infinite distances. We scale the whole scene to make the reconstructed scene inside as many Unit spheres as possible. This scaling is a hyperparameter that needs to be considered in conjunction with the values of the SFM reconstruction and the actual size of the scene. Our tests achieve the best results when each unit sphere contains the scene to be reconstructed. If the scalar is too large, i.e., the size of the Unit sphere is much larger than the scene, the reconstruction quality of will severely degrade the reconstruction quality of the close view. And if the scaling is too low, the overlap between the unit spheres will be reduced, which will result in a blurred image of the scene reconstruction like myopia. The space where we can perform novel view synthesis also needs to be covered by at least two unit spheres.

Actually, if we implicitly use rays to represent the scene, then the apporach above would be enough for our modeling section. However, there are problems with the original NeRF itself that cause it to not accurately model rays. Let us assume that there are two cameras in different positions and that the intersection of some two rays received by these two cameras. The sampling of that intersection point along the direction of these two rays should be obtained by querying the information of the pixel through which each of these two rays passes. In reality, however, the information of these two pixels represents two cones in space along the edges of the pixel from the camera point, not two rays of infinitely small diameter. Therefore the model can mistakenly take the information representing a finite volume as representing a definite point, which can cause the reconstruction of the area around that point to become blurred, as mentioned by mip-NeRF [6]. In addition, due to the presence of reprojection errors, this blurred area becomes larger as the distance of the sampled point coordinates from the scene center increases. This problem is much more severe for the panorama than for the standard perspective view because for the panorama projected onto the spherical model and the normal perspective image at the same resolution, the angular difference in the direction of the rays of adjacent pixels in the panorama is much larger than in the perspective view. The conical truncated body along the pixel edges of the panorama contains a larger volume than in the perspective view. As a result, the reconstruction quality will be very poor when we directly use the above approach to model the light of the panoramic image and apply the NeRF model. Also, the places farther away from the capturing position would become more blurred from the areas closer to the shooting point, no matter how we adjust the parameters and sampling methods. Therefore, we propose to use flexible spherical voxelization to represent the space.

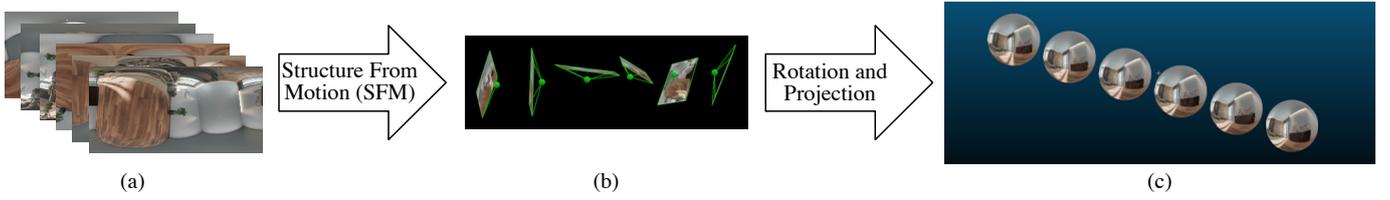

Fig. 3. Generation of training dataset: (a) Equirectangular images are captured by omnidirectional cameras with different pose and position; (b) camera position and orientation are acquired from SFM; (c) Equirectangular images are projected to Unit sphere model with the orientation and position obtained from (b). Therefore we get the ground truth of RGB information for different 5D input $(x, y, z, \theta, \phi)$

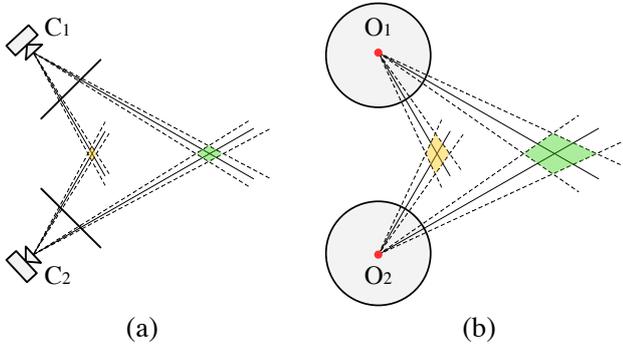

Fig. 4. Comparison of the sampling error of NeRF on perspective image and omnidirectional image: NeRF's point-based sampling ignore the shape and size viewed by each ray. This produces ambiguous point-sampled feature within the interception volume. In this image, we show and compare the interception volume size between the perspective and panorama views at the same resolution. It is obvious that omnidirectional camera has larger cross volume, and therefore the problem of ambiguous view is more severe, especially for distant areas.

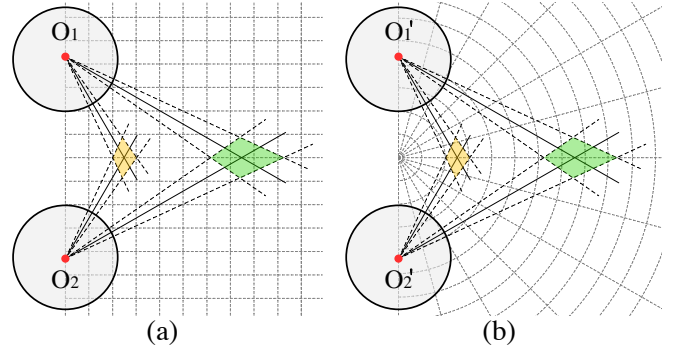

Fig. 5. The difference between cubic voxelization and spherical voxelization for panoramic cameras: As seen from the figure, the cross volume of rays sampled by panoramic cameras at different locations is different, and the use of spherical voxelization can effectively increase the overlapping between the voxels and the ray cross volume.

## VI. RAY SAMPLING ERROR AND VOXELIZATION

As we claimed in the main text, In NeRF [7], the ray is sampled by capturing the pixel through which the light passes, and the ray information represented by the pixel includes not only the ray but also a volume of the cone-truncated head that passes through the pixel from the camera, which can lead to blurring when generating a new viewpoint due to reprojection errors [6]. As shown in Fig. 4, the above ambiguity of the omnidirectional images is much higher than perspective ones. Therefore, voxelization method could improve the overall quality of the reconstruction.

However, since the density of intersection points between panoramic cameras becomes sparse as the distance from the filmed scene increases, a uniform voxelization method will result in different amounts of light passing through different voxels, which will reduce the efficiency of the model in learning the overall scene. Thus we try to adopt a spherical voxel method as shown in Fig. 5 to alleviate this problem and achieve some indoor-scenes results.

## VII. DATASET DETAIL

We tested our method on the dataset made by our own. The components of this dataset are collected from different sources, including generated images rendered by Blender and natural images of multiple scenes captured by commercially available omnidirectional cameras. For synthesized data, we acquire camera parameters from Blender during rendering. For actual data, we use Pix4D mapper software to get the extrinsic parameters of the cameras. The number of synthesized images is 1000 in total with the primitive resolution of $1024 \times 2048$.

The natural dataset consists of indoor and outdoor scenes. The indoor scene comprises 315 images captured by a tripod-mounted Insta360 OneX camera with a native resolution of $3040 \times 6080$. The cameras are placed $7 \times 9 \times 5$ cubic lattices in the capturing space. It is noteworthy that these images are captured under static lighting conditions without any disturbance around. Also, this dataset contains very complex geometry and is very challenging for the omnidirectional novel view synthesis task. We also provide the ground truth point cloud from 3 points in this space captured by Leica BLK360 LiDAR scanner for point cloud registration and depth estimation tasks. We hope our dataset will help to promote the research for computer vision on omnidirectional images.

The images in both outdoor scenes datasets are obtained from a tripod-mounted Theta S camera by hand-holding video and sampling at 30 frames per second at a frame rate with a native resolution of $960 \times 1920$. All images and camera parameters of the dataset will be released at the same time as the paper publication; however, only the indoor dataset contains the ground truth point cloud captured by the LiDAR scanner.

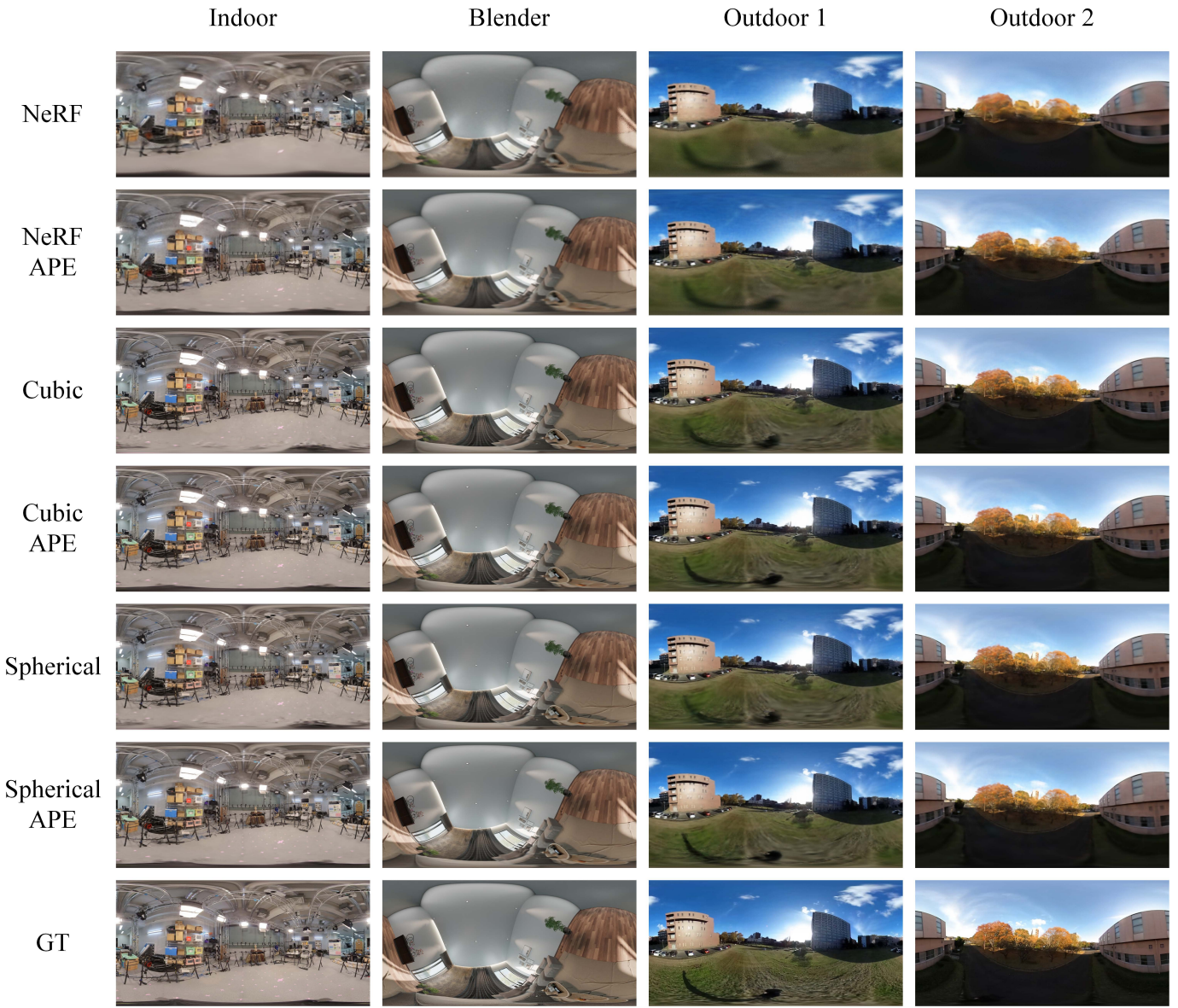

Fig. 6. Full images used in the main text to compare the experimental results

## VIII. Comparison Image

We attached the full image for comparison in the next page. As a result, our voxelization method increased the overall reconstruction quality holistically.



\end{document}